\documentclass[runningheads]{llncs}
\usepackage[T1]{fontenc}
\usepackage{graphicx,verbatim}
\usepackage{algorithm}

\usepackage{algpseudocode}

\usepackage{amsmath}
\usepackage{amssymb}
\usepackage{booktabs}
\usepackage{colortbl}%
\usepackage{subcaption}
\usepackage{hyperref}
\newsavebox{\rightblock}
\newcommand{\myrowcolor}{\rowcolor[gray]{0.925}}
\begin{document}
\title{MPFlow: Multi-modal Posterior-Guided Flow Matching for Zero-Shot MRI Reconstruction}
%

\author{
Seunghoi Kim$^{1,2}$\thanks{Correspondence to: seunghoi.kim.17@ucl.ac.uk} \:
Chen Jin$^{3}$ \,
Henry F.~J.~Tregidgo$^{1}$ \,
Matteo Figini$^{1}$ \,
Daniel C.~Alexander$^{1}$ \,
}

\authorrunning{Kim et al.}
\institute{
$^{1}$ University College London, UK \quad
$^{2}$ Heartflow, Inc., Mountain View, CA, USA \\
$^{3}$ Centre for AI, DS\&AI, AstraZeneca, UK 
}
  
\maketitle              
\begin{abstract}
Zero-shot MRI reconstruction relies on generative priors, but single-modality unconditional priors produce hallucinations under severe ill-posedness. 
In many clinical workflows, complementary MRI acquisitions (e.g. high-quality structural scans) are routinely available, yet existing reconstruction methods lack mechanisms to leverage this additional information.
We propose \emph{MPFlow}, a zero-shot multi-modal reconstruction framework built on rectified flow that incorporates auxiliary MRI modalities at inference time without retraining the generative prior to improve anatomical fidelity.
Cross-modal guidance is enabled by our proposed self-supervised pretraining strategy, \emph{Patch-level Multi-modal MR Image Pretraining (PAMRI)}, which learns shared representations across modalities.
Sampling is jointly guided by data consistency and cross-modal feature alignment using pre-trained \emph{PAMRI}, systematically suppressing intrinsic and extrinsic hallucinations.
Extensive experiments on HCP and BraTS show that MPFlow matches diffusion baselines on image quality using only 20\% of sampling steps while reducing tumor hallucinations by more than 15\% (segmentation dice score). This demonstrates that cross-modal guidance enables more reliable and efficient zero-shot MRI reconstruction. Code is available at \href{https://github.com/edshkim98/MPFlow}{https://github.com/edshkim98/MPFlow}.


\end{abstract}

\section{Introduction}
Magnetic resonance imaging (MRI) reconstruction from low-quality measurements is an important inverse problem in medical imaging. 
MRI data are often sub-sampled or acquired with thick slices to reduce acquisition time or improve signal-to-noise ratio, resulting in substantial information loss.

Recent deep learning methods have achieved strong performance in MRI reconstruction~\cite{iqt_pio,synth_survey,synthsr,diffusioniqt,iqt_stochastic,sr_mrigan,e2evarnet,sr_soupgan}, with diffusion models emerging as powerful generative priors for inverse problems in medical imaging. These diffusion-based approaches have been extended to zero-shot reconstruction~\cite{chung2023diffusion,ddrm,zeroddrm,DynamicDPS,zeroshotmri}, where a learned prior is guided by data consistency during sampling, eliminating the need for paired supervision. More recently, flow matching has been explored to improve sampling efficiency within the natural image domain~\cite{flowchef,flowdps}. However, these methods rely on unconditional priors, which may produce anatomically plausible yet incorrect reconstructions under severe ill-posedness. 

This limitation manifests as \emph{hallucinations}, which can be categorized as \emph{intrinsic}, violating measurement consistency, or \emph{extrinsic}, remaining measurement-consistent but unsupported by the ground truth (e.g. within the measurement null space)~\cite{hallucination_tomo,DynamicDPS}. 
Importantly, extrinsic hallucinations may persist even when data-consistency errors are small. 
Single-modality priors lack a mechanism to resolve ambiguity among multiple plausible null-space solutions without updating its prior or providing additional information. 
In clinical practice, multi-sequence protocols, such as multi-parametric imaging (diffusion, perfusion MRI) or multi-contrast structural scans, are routinely acquired and provide complementary anatomical information.
Although conditional approaches have explored multi-modal MRI reconstruction~\cite{tesla,3dmulticontrastcnn,multicontrasttransformer,alp}, they do not offer a principled mechanism to incorporate auxiliary modalities when the generative prior is unconditional.

\begin{figure}[t]
\centering
\includegraphics[width=\textwidth]{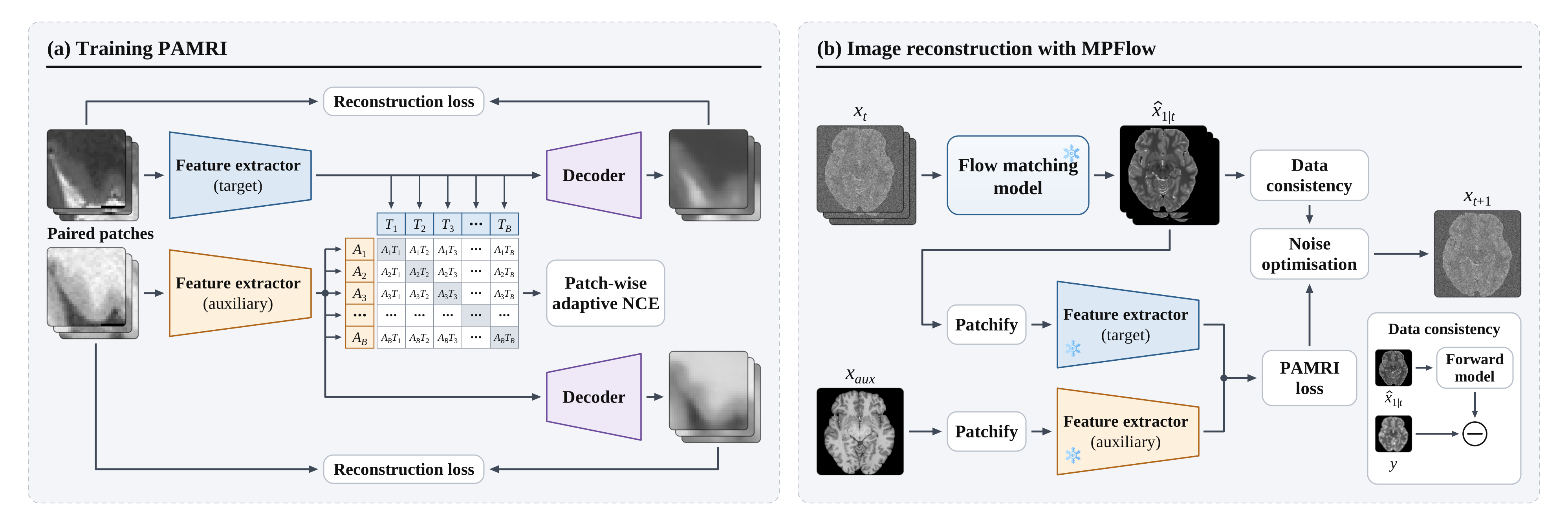}
\caption{\textbf{Schematic diagram of the proposed framework.} (a) \emph{PAMRI}: cross-modal feature alignment by learning a shared representation space across different MRI contrasts, (b) \emph{MPFlow}: flow-matching prior is guided by both data consistency and \emph{PAMRI}, to reduce intrinsic and extrinsic hallucinations.}
\label{fig:PAMRI}
\end{figure}

To address this limitation, we introduce \emph{MPFlow}, a zero-shot multi-modal MRI reconstruction framework based on flow matching that incorporates auxiliary MRI acquisitions at inference time without updating its unconditional prior. As illustrated in Fig.~\ref{fig:PAMRI}, we first train \emph{PAMRI}, a self-supervised framework that learns patch-level cross-modal correspondences between target (e.g. T2) and auxiliary (e.g. T1) modalities via a patch-wise adaptive InfoNCE loss. 
The temperature is dynamically chosen by normalized mutual information of paired patches to relax the contrastive penalty for augmentation-distorted positive pairs, preserving modality-specific structural details. At inference, the rectified flow prior is jointly guided by data consistency and \emph{PAMRI}, reducing \emph{intrinsic and extrinsic hallucinations}. To the best of our knowledge, this is the first zero-shot MRI reconstruction framework incorporating auxiliary imaging modalities at inference time.

Our contributions are threefold:
(1) We formulate multi-modal zero-shot MRI reconstruction, where an unconditional prior leverages auxiliary modalities at inference without modifying the prior, and show theoretically and empirically that such conditioning reduces hallucinations.
(2) We propose \emph{MPFlow}, a zero-shot multi-modal flow-matching framework, integrating \emph{PAMRI}, our self-supervised patch-level cross-modal alignment strategy, into a posterior update with noise optimization to mitigate poor trajectory initialization.
(3) We demonstrate that MPFlow matches diffusion baselines in image quality with only 20\% of the sampling steps, while reducing hallucinations by 15\% in tumor segmentation (Dice score) and 26\% in hallucination score (SHAFE) on BraTS and HCP.

\section{Method}

\subsection{Problem Formulation}

Let $\mathcal{X} \in \mathbb{R}^N$ for spatial domain tasks (e.g. super-resolution) 
and $\mathcal{X} \in \mathbb{C}^N$ for k-space tasks (e.g. subsampling), 
and let $x_{\mathrm{true}} \in \mathcal{X}$ denote the target high-quality MRI scan.
We consider the general degradation model:
\begin{equation}
    y = \mathcal{F}(x_{\mathrm{true}}) + \boldsymbol{\eta},
\end{equation}
where $y \in \mathcal{Y}$ are the acquired measurements, 
$\boldsymbol{\eta} \sim \mathcal{N}(0, \sigma^2 I)$ is additive Gaussian measurement noise, and 
$\mathcal{F} : \mathcal{X} \rightarrow \mathcal{Y}$
is the forward operator (e.g. k-space subsampling, spatial downsampling, or blurring).
Generative priors such as diffusion or flow models regularize reconstruction via posterior inference,
$\hat{x} \sim p(x \mid y)$, 
combining data consistency with a learned prior. 
However, when $\mathcal{F}$ is highly ill-posed, the posterior may assign probability mass to anatomically plausible but incorrect solutions, leading to hallucinations. 
While data consistency reduces intrinsic hallucinations, extrinsic hallucinations can still persist without prior-update or injecting additional information.

\subsection{Hypothesis and Justification}
We hypothesize that when registered modalities encode the same anatomy with complementary information, joint conditioning on measurements and an auxiliary modality reduces both intrinsic and extrinsic hallucinations.

\noindent\textbf{Justification}
Let $x$ denote the target-modality image, and $x_{\mathrm{aux}}$ a high-quality auxiliary modality. In a unimodal setting, the posterior $p(x \mid y) \propto p(y \mid x)\,p(x)$ is typically broad due to the weakly informative likelihood (under highly ill-posed setting), leading to extrinsic hallucinations.

From an information-theoretic perspective, the remaining uncertainty is quantified by the conditional entropy \(H(x \mid y)\). When additionally conditioning on an auxiliary modality, the posterior becomes
\(
p(x \mid y, x_{\mathrm{aux}}) \propto p(y \mid x)\,p(x \mid x_{\mathrm{aux}}),\)
and the uncertainty satisfies:
\begin{equation}
H(x \mid y, x_{\mathrm{aux}}) = H(x \mid y) - I(x; x_{\mathrm{aux}} \mid y),
\end{equation}
where \(I(x; x_{\mathrm{aux}} \mid y)\) denotes the conditional mutual information provided by the auxiliary modality beyond the measurement. Since registered MR modalities share overlapping but non-identical anatomical information, \(I(x; x_{\mathrm{aux}} \mid y) > 0\), implying \(H(x \mid y, x_{\mathrm{aux}}) < H(x \mid y)\).
Thus, data consistency and auxiliary conditioning can suppress intrinsic and extrinsic hallucinations, respectively. 

\subsection{Rectified Flow} 
We use rectified flow~\cite{rectifiedflow} from flow-matching as the generative prior.
Rectified flow defines a continuous-time probability flow from a simple base distribution
(e.g. Gaussian noise) to the data distribution via the ODE:
\begin{equation}
\label{eq:fm_ode}
    \frac{d x_t}{d t} = v_\theta(x_t,t), \quad t \in [0,1],
\end{equation}
where $v_\theta$ is a learned velocity field.
We adopt a straight-line interpolation between noise
$z \sim \mathcal{N}(0,I)$ at $t=0$ and data $x_1$ at $t=1$:
\begin{equation}
\label{eq:linear_path}
    x_t = (1-t) z + t x_1,
    \qquad
    v(x_t,t) = \frac{d x_t}{d t} = x_1 - z.
\end{equation}
The rectified flow objective minimizes
\begin{equation}
\label{eq:fm_loss}
    \mathcal{L}_{\mathrm{FM}} =
    \mathbb{E}_{t,z,x_1}
    \big[\|v_\theta(x_t,t) - (x_1 - z)\|_2^2\big].
\end{equation}
Through this objective, rectified flow learns to approximate linear transport from noise to data. The near-linear and deterministic trajectories enable high-quality image generation with substantially fewer sampling steps than diffusion models.


\subsection{Patch-level Multi-modal MR Image Pretraining (PAMRI)}
\label{subsec:pamri}

We establish \emph{PAMRI}, a contrastive feature alignment framework for multi-modal MR images. \emph{PAMRI} uses independent encoders, $\phi(\cdot)$ for the target and $\psi(\cdot)$ for the auxiliary modality, to map them into a shared latent space. This independence disentangles modality-specific appearance from shared anatomical semantics. We adopt a lightweight yet effective ResNet18~\cite{resnet} architecture for both encoders to minimize computational overhead at inference.

MRI reconstruction is a dense task requiring fine-grained structural information, but conventional contrastive learning discards spatial detail. 
To preserve it, we operate on patches (e.g. $32\times32$ pixels) and introduce an auxiliary reconstruction task.
Given a batch of $B$ paired patches $\{(p_i^{\mathrm{tar}},p_i^{\mathrm{aux}})\}_{i=1}^B$, we compute normalized latent embeddings $u_i = \phi(p_i^{\mathrm{tar}})/\|\phi(p_i^{\mathrm{tar}})\|_2$ and $w_i = \psi(p_i^{\mathrm{aux}})/\|\psi(p_i^{\mathrm{aux}})\|_2$. 
This alignment is driven by our proposed adaptive contrastive objective. Building upon InfoNCE loss~\cite{simclr}, our loss evaluates both inter- and intra-modal negatives: $\mathcal{L}_{\mathrm{NCE}} = \frac{1}{2B}\sum_{i=1}^B \left( \ell(u_i, w_i) + \ell(w_i, u_i) \right)$. 
The loss for a specific positive pair $(e_i, e_j)$, where $e_i \in \{u_i, w_i\}$, is defined as:
\begin{equation}
\ell(e_i, e_j) = 
-\log 
\frac{\exp(e_i^\top e_j/\tau_{i,j})}
{\sum_{k=1}^B \exp(e_i^\top u_k/\tau_{i,k}) 
+ \sum_{k=1}^B \exp(e_i^\top w_k/\tau_{i,k})
- \exp(e_i^\top e_i/\tau_{i,i})}.
\end{equation}
Here, $\tau_{i,j}$ is an adaptive temperature scaled by the Normalized Mutual Information (NMI) between their corresponding raw patches: $\tau_{i,j} = \tau_{\min} + (\tau_{\max} - \tau_{\min})(1 - \mathrm{NMI}(p_i, p_j))$. 
This dynamically adjusts the contrastive penalty based on the inherent multi-modal information overlap. Specifically, for positive pairs substantially distorted by augmentations, the temperature increases, relaxing the pull to preserve modality-specific structural 
details crucial for downstream dense reconstruction.

Lightweight decoders $d_{\mathrm{tar}}, d_{\mathrm{aux}}$ are attached to regularize the representations via patch reconstruction:
\begin{equation}
\label{eq:rec_loss}
\mathcal{L}_{\mathrm{rec}} = 
\frac{1}{B}\sum_{i=1}^B 
\Big(
\|d_{\mathrm{tar}}(\phi(p_{i}^{\mathrm{tar}}))-p_i^{\mathrm{tar}}\|_1 
+ 
\|d_{\mathrm{aux}}(\psi(p_i^{\mathrm{aux}}))-p_i^{\mathrm{aux}}\|_1
\Big).
\end{equation}

The final objective becomes $\mathcal{L}_{\mathrm{SSL}} = \mathcal{L}_{\mathrm{NCE}} + \lambda_{\mathrm{rec}}\mathcal{L}_{\mathrm{rec}}$.
Rather than using mutual information as a direct pixel-level alignment loss, which struggles with multi-modal intensity variations (e.g. a tumor is bright in FLAIR but nearly isointense in T1), \emph{PAMRI} elegantly uses it to weight the latent contrastive space. By dynamically adjusting the temperature for naturally dissimilar patches, \emph{PAMRI} learns robust, modality-invariant representations that enable spatially structured and semantically meaningful posterior guidance.

\subsection{Multi-modal Posterior-Guided Flow Matching}
We perform posterior inference by guiding rectified flow with data consistency (DC) and cross-modal alignment using pre-trained \emph{PAMRI} encoders from Sec.~\ref{subsec:pamri}. At each timestep $t$, we project the current state to the clean image manifold via the estimate \( \hat{x}_{1|t} = x_t - (1-t)\, v_\theta(x_t,t)\) to drive the posterior update.
Over $T$ sampling steps, the update in each time step becomes:
\begin{equation}
\label{eq:mpflow_update}
v(x_t|y) 
= \underbrace{v_\theta(x_t,t)}_{\text{Prior}} 
- \alpha_t \nabla_{x_t}
\Bigg(
\underbrace{ \|\mathcal{F}(\hat{x}_{1|t})-y\|_2^2}_{\text{Intrinsic Reduction}}
+ 
\underbrace{\lambda_{\text{P}} \mathcal{L}_{\text{P}}(\phi(\mathcal{P}(\hat{x}_{1|t})), \psi(\mathcal{P}(x_{\text{aux}})))}_{\text{Extrinsic Reduction}}
\Bigg),
\end{equation}

where \(v_\theta(x_t,t)\) and \(\alpha_t\) denote the prior drift and step size. \(\lambda_{\mathrm{P}}\) controls the $\mathcal{L}_{\text{P}}$'s strength. 
$\mathcal{L}_{\text{P}}$ is MSE between normalized latent features of patches extracted (excluding background) via $\mathcal{P}(\cdot)$ from the current estimate $\hat{x}_{1|t}$ and the auxiliary $x_{\text{aux}}$.
DC reduces measurement space deviation, and $\mathcal{L}_{\text{P}}$ minimizes deviations from the auxiliary image in the latent space, systematically tackling both intrinsic and extrinsic hallucinations.

\subsubsection{Initial noise optimization}
The initial noise $z$ can influence the reconstruction quality of flow-matching. To mitigate poor initializations, we sample $S$ candidate seeds $\{z^{(s)}\}_{s=1}^S \sim \mathcal{N}(0,I)$ and perform a short warm-start of $t_{\mathrm{noise}}$ steps in parallel. Each candidate is evaluated using the composite objective
\(
\Phi(x) = \|\mathcal{F}(x)-y\|_2^2 
+ \lambda_{\mathrm{P}} \mathcal{L}_{\mathrm{P}}(x, x_{\mathrm{aux}}),
\)
and we select:
\begin{equation}
\label{eq:noise_opt}
s^\star = \arg\min_{s \in \{1,\dots,S\}} 
\Phi\!\left(\hat{x}^{(s)}_{1|t_{\mathrm{noise}}}\right).
\end{equation}
Sampling then continues only with the selected seed. Unlike previous method~\cite{dpps} that optimize via DC only, $\Phi(\cdot)$ jointly uses DC and \emph{PAMRI} to select the candidate with lowest intrinsic and extrinsic hallucinations, yielding better posterior trajectories with minimal overhead.
\section{Experiments}
\subsection{Experimental Setup}
\subsubsection{Datasets}
We evaluate our method on two widely used datasets: the Human Connectome Project (HCP)~\cite{hcp} and BraTS~\cite{brats}. We perform T2 super-resolution ($4\times$) on HCP, and FLAIR k-space reconstruction ($8\times$) on BraTS, both using fully sampled T1 as the auxiliary modality. For HCP, we follow prior work~\cite{DynamicDPS} for the testing (N=200) and for BraTS, we use the official validation set (N=250) to follow the standard protocol. Experiments are conducted on axial slices.
\subsubsection{Implementation Details}
Diffusion and rectified flows are trained on same dataset for 100k iterations. 
For \emph{MPFlow}, we set $\lambda_{\text{P}} = 0.1$, $\lambda_{\text{rec}} = 0.5$, $S = 8$, and $t_{\text{noise}} = 0.2T$. Patch size $32 \times 32$ is used for \emph{PAMRI} pretraining. 
\subsubsection{Baselines \& Metrics}
We compare against DIP~\cite{dip}, DPS~\cite{chung2023diffusion}, DiffDeuR~\cite{zeroshotmri}, and DynamicDPS~\cite{DynamicDPS} (without warm-start). 
Since no existing zero-shot method incorporates auxiliary imaging modalities, all baselines are unimodal.
Reconstruction quality is evaluated using PSNR, SSIM, and LPIPS~\cite{lpips}. 
Hallucinations are assessed via (i) measurement-space error, (ii) tumor segmentation Dice score, and (iii) SHAFE (Semantic Hallucination Assessment via Feature Evaluation)~\cite{shafe}. 
For Dice score, a Swin-UNet~\cite{swinunet} trained on BraTS FLAIR is applied to both ground-truth and reconstructed images. SHAFE measures semantic feature discrepancies via a pre-trained vision encoder (DINO~\cite{dinov3}) with exponentially weighted aggregation to emphasize hallucinated structures.
\begin{table}[!htbp]
  \centering
  \footnotesize
  \caption{\textbf{Quantitative comparisons on HCP (SR $\times$4) and BraTS (Acc. $\times$8).}
  $T$ denotes total time steps; DDIM used for $T=100$.
  p-values are calculated against the second-best baseline for each metric.}
  \renewcommand{\arraystretch}{1.1}
  \begin{tabular}{c c ccc ccc}
    \toprule
      & & \multicolumn{3}{c}{HCP (SR $\times$4)} 
        & \multicolumn{3}{c}{BraTS (Acc. $\times$8)} \\
      \cmidrule(lr){3-5} \cmidrule(lr){6-8}
      \textbf{Method} 
        & $T$
        & \textit{PSNR} ($\uparrow$) 
        & \textit{SSIM} ($\uparrow$) 
        & \textit{LPIPS} ($\downarrow$) 
        & \textit{PSNR} ($\uparrow$) 
        & \textit{SSIM} ($\uparrow$) 
        & \textit{LPIPS} ($\downarrow$) \\
    \midrule
    DIP~\cite{dip} 
      & 3K 
      & 24$\pm$.7 & .66$\pm$.03 & .30$\pm$.03 
      & 25$\pm$2.4 & .57$\pm$.10 & .19$\pm$.01 \\
    DiffDeuR~\cite{zeroshotmri} 
      & .5K
      & 23$\pm$1.6 & .78$\pm$.04 & .23$\pm$.04 
      & 30$\pm$1.9 & .88$\pm$.02 & .12$\pm$.02 \\
    \myrowcolor
    DPS~\cite{chung2023diffusion} 
      & .1K
      & 23$\pm$1.4 & .69$\pm$.16 & .20$\pm$.08 
      & 23$\pm$5.8 & .65$\pm$.27 & .23$\pm$.12 \\
    DPS~\cite{chung2023diffusion} 
      & .5K
      & 24$\pm$1.0 & .79$\pm$.03 & .15$\pm$.01 
      & 25$\pm$2.4 & .89$\pm$.01 & .09$\pm$.02 \\
    \myrowcolor
    DynamicDPS~\cite{DynamicDPS} 
      & .1K
      & 23$\pm$1.8 & .69$\pm$.18 & .21$\pm$.09 
      & 24$\pm$5.6 & .68$\pm$.16 & .22$\pm$.12 \\
    DynamicDPS~\cite{DynamicDPS} 
      & .5K
      & 24$\pm$.9 & .79$\pm$.03 & .15$\pm$.02 
      & 31$\pm$1.9 & .89$\pm$.02 & .09$\pm$.02 \\
    \midrule
    \myrowcolor
    MPFlow (Ours)
      & .1K
      & 23$\pm$.9 & .78$\pm$.03 & .16$\pm$.02 
      & \textbf{31}$\pm$1.8 & .90$\pm$.02 & .08$\pm$.01 \\
    MPFlow (Ours)
      & .5K
      & \textbf{25}$\pm$.9 & \textbf{.82}$\pm$.03 & \textbf{.14}$\pm$.02 
      & \textbf{31}$\pm$.9 & \textbf{.91}$\pm$.02 & \textbf{.07}$\pm$.01 \\
      \midrule
    p-value 
      & -
      & <.001 & <.001 & .031 
      & .016 & .001 & <.001 \\
    \bottomrule
  \end{tabular}
  \label{table:results}
\end{table}

\subsection{Main Results}
Tab.~\ref{table:results} presents quantitative comparisons under varying sampling budgets. 
Across all tasks and metrics, \emph{MPFlow} consistently outperforms zero-shot baselines, with all improvements statistically significant ($p<0.05$).
At $T=500$, \emph{MPFlow} outperforms the second best baseline (DynamicDPS) by 2--4\% in SSIM and 6--22\% in LPIPS, despite DynamicDPS requiring multiple inner steps per timestep, demonstrating \emph{MPFlow}'s efficiency. 
The improvements are more pronounced in the time-constrained setting ($T=100$).
Diffusion-based methods suffer large degradation with SSIM and LPIPS dropping 15--83\% and increased variance, indicating unstable inference due to large discretization errors from DDIM sampling. 
In contrast, \emph{MPFlow} exhibits marginal performance reduction (2.5--14\%), demonstrating superior efficiency and robustness as the near-linear trajectories of rectified flow permit larger step sizes without significant discretization error. 
Remarkably, \emph{MPFlow} matches the reconstruction quality of diffusion-based baselines at $T=500$ using only $T=100$ steps.

\begin{figure}[!htbp]
\centering
\includegraphics[width=\textwidth]{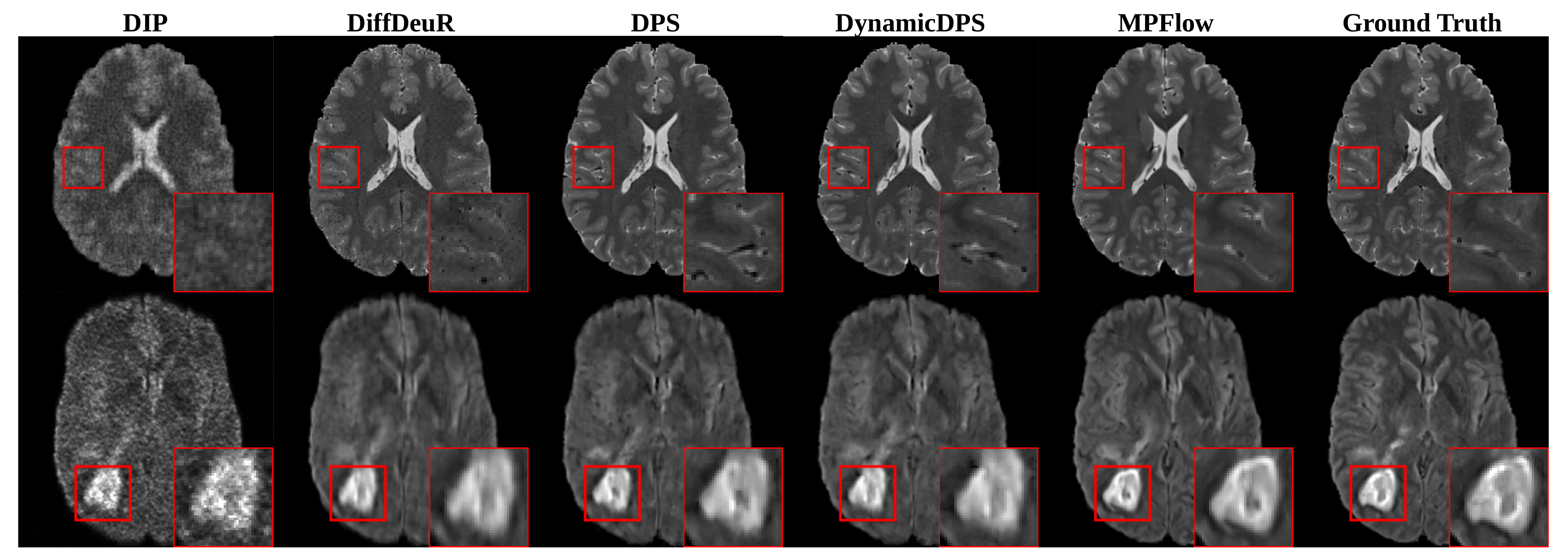}
\caption{\textbf{Visual comparisons on HCP (top), BraTS (bottom).} Highlighted red boxes show MPFlow clearly reduces hallucinations compared to the baselines.}
\label{fig:main_results}
\end{figure}

Fig.~\ref{fig:main_results} presents visual comparisons of \emph{MPFlow} against baseline methods. As highlighted in the red boxes, baselines introduce hallucinated structures including distorted sulci and incorrect tumor morphology, which can directly compromise clinical decisions such as surgical planning and radiotherapy contouring. In contrast, \emph{MPFlow} preserves anatomically faithful structures with notably sharper tumor boundaries, demonstrating the benefit of cross-modal alignment via \emph{PAMRI}.
These qualitative results further demonstrate \emph{MPFlow}'s ability to reduce hallucinations and improve structural fidelity in clinically critical regions.

\subsection{Further Analysis}
Global image quality metrics (e.g. PSNR, SSIM) often fail to capture hallucinations, as they prioritize global fidelity while overlooking sparse, localized failures~\cite{localdiff,shafe}.
Instead, we evaluate hallucinations in Tab.~\ref{tab:hallu_eval} using measurement-space loss, SHAFE, and tumor segmentation Dice score.
Across both datasets, \emph{MPFlow} substantially outperforms the baselines in all hallucination metrics. 
Across both datasets, \emph{MPFlow} (Full) substantially outperforms all baselines: it reduces measurement-space loss by over 80\% on HCP and 88\% on BraTS relative to DPS, while improving SHAFE by 38\% and Dice by 16\%, confirming that jointly guiding the prior with data consistency and \emph{PAMRI} suppresses both intrinsic and extrinsic hallucinations.
To isolate the contribution of each component, we ablate \emph{PAMRI} and noise optimization separately. Adding \emph{PAMRI} to the vanilla baseline (Base $\rightarrow$ PAMRI) yields a 27\% SHAFE reduction on HCP and 5\% Dice improvement on BraTS, with comparatively smaller measurement-loss change, demonstrating that cross-modal guidance primarily suppresses \emph{extrinsic} hallucinations by anchoring null-space content to the auxiliary anatomy. Conversely, noise optimization alone (Base $\rightarrow$ Noise) reduces measurement-space loss by 82\% on HCP and 89\% on BraTS with smaller SHAFE/Dice gains, indicating that better seed selection primarily suppresses \emph{intrinsic} hallucinations by initializing trajectories that converge to measurement-consistent solutions. This confirms that the two modules address complementary failure modes.


\begin{table}[t]
\centering
\footnotesize
\caption{\textbf{Quantitative hallucination evaluation on HCP and BraTS (T=100).} We evaluate hallucination across baselines using measurement-space loss, SHAFE (HCP) and dice score on tumor segmentation (BraTS).}
\label{tab:hallu_eval}
\begin{subtable}[t]{0.46\textwidth}
    \centering
    \footnotesize
    \caption{Hallucination Analysis (HCP)}
    \label{tab:hcp_eval}
    \begin{tabular}{lcc}
    \toprule
    \textbf{Method} & \textbf{Meas. L} $\downarrow$ & \textbf{SHAFE} $\downarrow$ \\
    \midrule
    DPS          & 3.84$\times 10^{-3}$ & 0.278 \\
    DynamicDPS   & 3.44$\times 10^{-3}$ & 0.274 \\
    \midrule
    MPFlow (Base) & 6.84$\times 10^{-3}$ & 0.233 \\
    MPFlow (PAMRI) & 1.43$\times 10^{-3}$ & 0.171 \\
    MPFlow (Noise) & \textbf{1.25$\times 10^{-3}$}  & 0.204  \\
    MPFlow (Full) & 1.30$\times 10^{-3}$ & \textbf{0.170} \\
    \bottomrule
    \end{tabular}
\end{subtable}
\hfill 
\begin{subtable}[t]{0.46\textwidth}
    \centering
    \footnotesize
    \caption{Tumor Segmentation (BraTS)}
    \label{tab:brats_eval}
    \begin{tabular}{lcc}
    \toprule
    \textbf{Method} & \textbf{Meas. L} $\downarrow$ & \textbf{Dice} $\uparrow$ \\
    \midrule
    DPS   & 4.20$\times 10^{-2}$ & 0.618 \\
    DynamicDPS   & 4.51$\times 10^{-2}$ & 0.639 \\
    \midrule
    MPFlow (Base) & 4.36$\times 10^{-3}$ & 0.704 \\
    MPFlow (PAMRI) & 2.45$\times 10^{-3}$ & 0.741 \\
    MPFlow (Noise) & 4.90$\times 10^{-4}$ & 0.735 \\
    MPFlow (Full) & \textbf{4.89$\times 10^{-4}$} & \textbf{0.743} \\
    \bottomrule
    \end{tabular}
\end{subtable}
\end{table}

Fig.~\ref{fig:pamri_comparison} compares PAMRI against simpler auxiliary guidance losses; normalized mutual information (NMI), Canny edge, and pixel-MSE. PAMRI outperforms these baselines on both image quality and hallucination score. Notably, Canny edge and pixel-MSE hallucinate more than the DC-only baseline, indicating that naive heuristic guidance can adversely affect reconstruction quality.
To assess \emph{PAMRI}'s role in reducing extrinsic hallucinations, we hypothesize that its benefit should scale with task severity, since greater ill-posedness enlarges the null-space ambiguity. Fig.~\ref{fig:task_difficulty} confirms this: $\Delta$SSIM grows from 3.81\% at 4$\times$ SR to 6.02\% at 8$\times$ SR, showing that cross-modal guidance becomes increasingly important as data consistency alone cannot resolve the null-space ambiguity.

\newlength{\imgheight}
\setlength{\imgheight}{3.5cm} 

\begin{figure}[h]
    \centering
    \footnotesize
    \caption{\textbf{Further analysis of MPFlow on HCP (T=100).} (a) PAMRI vs. intensity-based auxiliary losses on image quality (SSIM $\uparrow$) and hallucination (SHAFE $\downarrow$). (b) Effect of PAMRI under increasing super-resolution difficulty.}

    \begin{subfigure}[b]{0.48\textwidth}
        \centering
        \includegraphics[height=\imgheight]{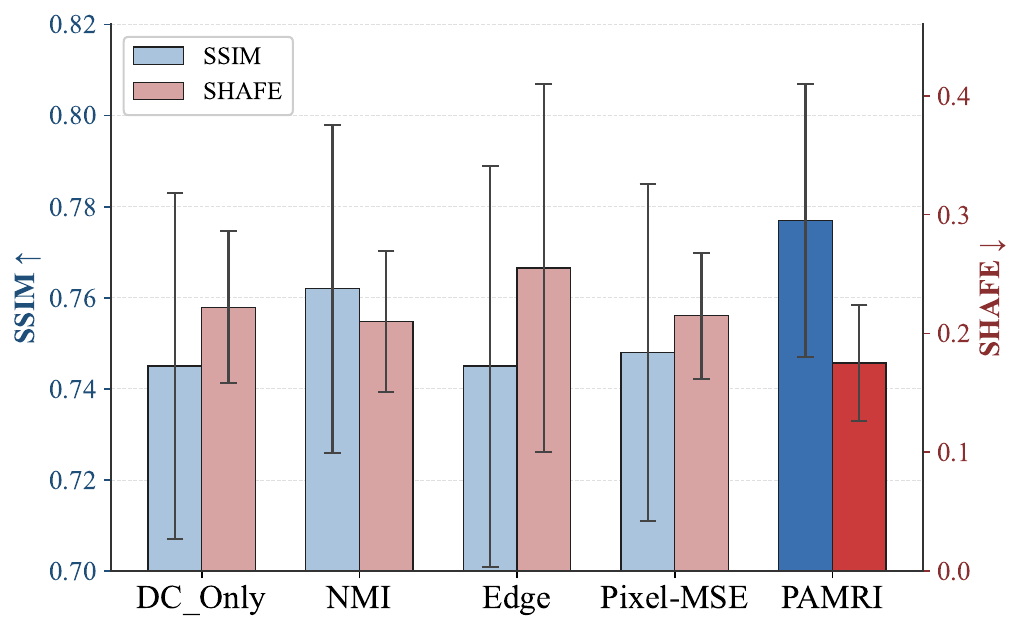}
        \caption{PAMRI yields the highest SSIM and lowest SHAFE}
        \label{fig:pamri_comparison}
    \end{subfigure}
    \hfill
    \begin{subfigure}[b]{0.48\textwidth}
        \centering
        \includegraphics[height=\imgheight]{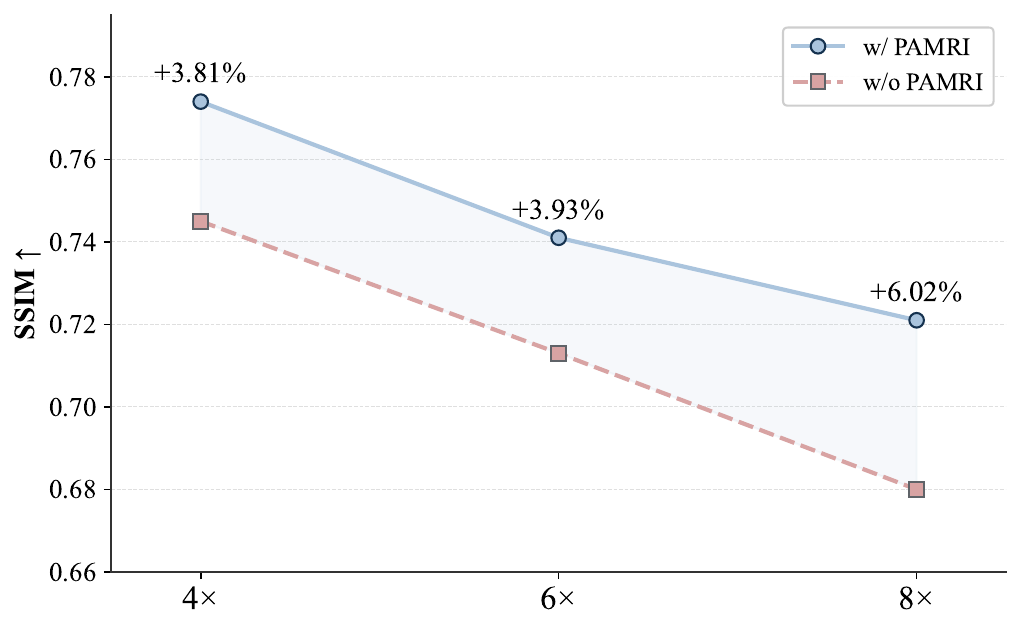}
        \caption{Benefit of PAMRI grows as super-resolution factor increases}
        \label{fig:task_difficulty}
    \end{subfigure}
\end{figure}

\section{Conclusion}
We propose \emph{MPFlow}, a multi-modal posterior-guided flow-matching framework for zero-shot MRI reconstruction. 
The auxiliary imaging modality is leveraged at inference time via \emph{PAMRI}, our self-supervised contrastive pretraining strategy that learns shared cross-modal representations. At inference, the unconditional prior is guided by data consistency and \emph{PAMRI}, suppressing both intrinsic and extrinsic hallucinations without updating the prior. Experiments on HCP and BraTS across super-resolution and k-space subsampling tasks show improvements in reconstruction fidelity and hallucination reduction. Future work explores extending the framework to non-imaging modalities and adaptive guidance schemes. In overall, our results show that auxiliary modalities can reshape posterior geometry at inference time under unconditional generative priors, enabling robust multi-modal inverse reconstruction without paired supervision.
\begin{credits}
\subsubsection{\ackname}
    This work is supported by the EPSRC-funded UCL Centre for Doctoral Training in Intelligent, Integrated Imaging in Healthcare (i4health) under grant number EP/S021930/1. 
    The work of DCA is supported by the Wellcome Trust awards 221915 and 317797, and EPSRC grant EP/Y028856/1. 
    The work of MF and HG is supported by the MRC (award MR/W031566/1), and the NIHR UCLH Biomedical Research Centre.
\subsubsection{\discintname} The authors have no competing interests to declare that are relevant to the content of this article.
\end{credits}

\bibliographystyle{splncs04}
\bibliography{main}

@String(CVPR= {IEEE Conf. Comput. Vis. Pattern Recog.})

@String(ICCV= {Int. Conf. Comput. Vis.})

@String(ECCV= {Eur. Conf. Comput. Vis.})

@String(CVPR  = {CVPR})

@String(ICCV  = {ICCV})

@String(ECCV  = {ECCV})

@inproceedings{ddrm,
    title={Denoising Diffusion Restoration Models},
    author={Bahjat Kawar and Michael Elad and Stefano Ermon and Jiaming Song},
    booktitle={Advances in Neural Information Processing Systems},
    year={2022}
}

@article{diffusioniqt,
  title={A 3D Conditional Diffusion Model for Image Quality Transfer--An Application to Low-Field MRI},
  author={Kim, Seunghoi and Tregidgo, Henry FJ and Eldaly, Ahmed K and Figini, Matteo and Alexander, Daniel C},
  journal={arXiv preprint arXiv:2311.06631},
  year={2023}
}

@article{zeroddrm,
title={Zero-Shot Image Restoration Using Denoising Diffusion Null-Space Model},
author={Wang, Yinhuai and Yu, Jiwen and Zhang, Jian},
journal={The Eleventh International Conference on Learning Representations},
year={2023}
}

@InProceedings{localdiff,
author="Kim, Seunghoi
and Jin, Chen
and Diethe, Tom
and Figini, Matteo
and Tregidgo, Henry F. J.
and others",
title="Tackling Structural Hallucination in Image Translation with Local Diffusion",
booktitle="European Conference on Computer Vision (ECCV)",
year="2024",
}

@inproceedings{
chung2023diffusion,
title={Diffusion Posterior Sampling for General Noisy Inverse Problems},
author={Hyungjin Chung and Jeongsol Kim and Michael Thompson Mccann and Marc Louis Klasky and Jong Chul Ye},
booktitle={The Eleventh International Conference on Learning Representations },
year={2023},
}

@article{dip,
    author    = {Ulyanov, Dmitry and Vedaldi, Andrea and Lempitsky, Victor},
    title     = {Deep Image Prior},
    journal   = {Proceedings of the IEEE/CVF Conference on Computer Vision and Pattern Recognition},
    year      = {2018}
}

@ARTICLE{brats,
  author={Menze, Bjoern H. and Jakab, Andras and Bauer, Stefan and Kalpathy-Cramer, Jayashree and Farahani, Keyvan et al.},
  journal={IEEE Transactions on Medical Imaging}, 
  title={The Multimodal Brain Tumor Image Segmentation Benchmark (BRATS)}, 
  year={2015},
  volume={34},
  number={10},
  pages={1993-2024},
}

@article{hcp,
  title={Advances in diffusion MRI acquisition and processing in the Human Connectome Project},
  author={Stamatios N. Sotiropoulos and Sa{\^a}d Jbabdi and Junqian Xu and Jesper L. R. Andersson and Steen Moeller and others},
  journal={NeuroImage},
  year={2013},
  volume={80},
  pages={125-143}
}

@InProceedings{zeroshotmri,
        author = {Lin, Xiyue and Du, Chenhe and Wu, Qing and Tian, Xuanyu and Yu, Jingyi and others},
        title = { { Zero-shot Low-field MRI Enhancement via Denoising Diffusion Driven Neural Representation } },
        booktitle = {proceedings of Medical Image Computing and Computer Assisted Intervention -- MICCAI 2024},
        year = {2024}
}

@article{iqt_pio,
    author = {Alexander, Daniel C. and Darko, Zikic and Ghosh,  Aurobrata and Tanno, Ryutaro and Wottschel, Viktor and others},
    title = {Image quality transfer and applications in diffusion MRI},
    journal = {NeuroImage},
    year = {2017},
    volume = {152},
    pages = {283--298},
}

@article{iqt_stochastic,
  title = {Low-field magnetic resonance image enhancement via stochastic image quality transfer}, 
  journal = {Medical Image Analysis},
  volume = {87},
  pages = {102807},
  year = {2023},
  issn = {1361-8415},
  author = {Hongxiang Lin and Matteo Figini and Felice D'Arco and Godwin Ogbole and Ryutaro Tanno and others},
}

@article{synthsr,
  author = {Juan E. Iglesias  and Benjamin Billot  and Yaël Balbastre  and Colin Magdamo  and Steven E. Arnold and others},
  title = {SynthSR: A public AI tool to turn heterogeneous clinical brain scans into high-resolution T1-weighted images for 3D morphometry},
  journal = {Science Advances},
  volume = {9},
  number = {5},
  year = {2023},
}

@InProceedings{DynamicDPS,
        author = { Kim, Seunghoi AND Tregidgo, Henry F. J. AND Figini, Matteo AND Jin, Chen AND Joshi, Sarang AND Alexander, Daniel C.},
        title = { { Tackling Hallucination from Conditional Models for Medical Image Reconstruction with DynamicDPS } },
        booktitle = {proceedings of Medical Image Computing and Computer Assisted Intervention -- MICCAI 2025},
        year = {2025}
}

@article{hallucination_tomo,
  author    = {Bhadra, S. and Kelkar, V. A. and Brooks, F. J. and Anastasio, M. A.},
  title     = {On Hallucinations in Tomographic Image Reconstruction},
  journal   = {IEEE Transactions on Medical Imaging},
  year      = {2021},
  volume    = {40},
  number    = {11},
  pages     = {3249--3260},
}

@inproceedings{lpips,
  title={The Unreasonable Effectiveness of Deep Features as a Perceptual Metric},
  author={Zhang, Richard and Isola, Phillip and Efros, Alexei A and Shechtman, Eli and Wang, Oliver},
  booktitle={Proceedings of the IEEE/CVF Conference on Computer Vision and Pattern Recognition (CVPR)},
  year={2018}
}

@article{synth_survey,
    author = {Gopinath, Karthik and Hoopes, Andrew and Alexander, Daniel C. and Arnold, Steven E. and Balbastre, Yael and others},
    title = {Synthetic data in generalizable, learning-based neuroimaging},
    journal = {Imaging Neuroscience},
    volume = {2},
    pages = {1-22},
    year = {2024},
}

@article{sr_soupgan,
  title={SOUP-GAN: Super-Resolution MRI Using Generative Adversarial Networks},
  author={Kuan Zhang and Haoji Hu and Kenneth A. Philbrick and Gian Marco Conte and Joseph D. Sobek and others},
  journal={Tomography},
  year={2021},
  volume={8},
  pages={905 - 919}
}

@inproceedings{sr_mrigan,
    author = {Wang, Jiancong and Chen, Yuhua and Wu, Yifan and Shi, Jianbo and Gee, James},
    title = {Enhanced generative adversarial network for 3D brain MRI super-resolution},
    booktitle = {IEEE/CVF Winter Conference (WACV)},
    year = {2020},
    location = {Conference Location},
}

@misc{dinov3,
  title={{DINOv3}},
  author={Sim{\'e}oni, Oriane and Vo, Huy V. and Seitzer, Maximilian and Baldassarre, Federico and Oquab, Maxime and Jose, Cijo and Khalidov, Vasil and Szafraniec, Marc and Yi, Seungeun and Ramamonjisoa, Micha{\"e}l and Massa, Francisco and Haziza, Daniel and Wehrstedt, Luca and Wang, Jianyuan and Darcet, Timoth{\'e}e and Moutakanni, Th{\'e}o and Sentana, Leonel and Roberts, Claire and Vedaldi, Andrea and Tolan, Jamie and Brandt, John and Couprie, Camille and Mairal, Julien and J{\'e}gou, Herv{\'e} and Labatut, Patrick and Bojanowski, Piotr},
  year={2025},
  eprint={2508.10104},
  archivePrefix={arXiv},
}

@article{e2evarnet,
  title={End-to-End Variational Networks for Accelerated MRI Reconstruction},
  author={Anuroop Sriram and Jure Zbontar and Tullie Murrell and Aaron Defazio and C. Lawrence Zitnick and Nafissa Yakubova and Florian Knoll and Patricia M. Johnson},
  journal={ArXiv},
  year={2020}
}

@InProceedings{tesla,
        author = { Choi, Yoonseok AND Jung, Sunyoung AND Al-masni, Mohammed A. AND Yang, Ming-Hsuan AND Kim, Dong-Hyun},
        title = { { TESLA: Test-time Reference-free Through-plane Super-resolution for Multi-contrast Brain MRI } },
        booktitle = {proceedings of Medical Image Computing and Computer Assisted Intervention -- MICCAI 2025},
        year = {2025},
}

@article{3dmulticontrastcnn,
title = {3D-MRI super-resolution reconstruction using multi-modality based on multi-resolution CNN},
journal = {Computer Methods and Programs in Biomedicine},
volume = {248},
pages = {108110},
year = {2024},
author = {Li Kang and Bin Tang and Jianjun Huang and Jianping Li},
}

@INPROCEEDINGS{multicontrasttransformer,
  author={Li, Guangyuan and Zhao, Lei and Sun, Jiakai and Lan, Zehua and Zhang, Zhanjie and Chen, Jiafu and Lin, Zhijie and Lin, Huaizhong and Xing, Wei},
  booktitle={2023 IEEE/CVF International Conference on Computer Vision (ICCV)}, 
  title={Rethinking Multi-Contrast MRI Super-Resolution: Rectangle-Window Cross-Attention Transformer and Arbitrary-Scale Upsampling}, 
  year={2023},
}

@misc{shafe,
      title={HalluGen: Synthesizing Realistic and Controllable Hallucinations for Evaluating Image Restoration}, 
      author={Seunghoi Kim and Henry F. J. Tregidgo and Chen Jin and Matteo Figini and Daniel C. Alexander},
      year={2025},
      eprint={2512.03345},
      archivePrefix={arXiv},
      primaryClass={cs.CV},
      url={https://arxiv.org/abs/2512.03345}, 
}

@article{simclr,
  title={A Simple Framework for Contrastive Learning of Visual Representations},
  author={Chen, Ting and Kornblith, Simon and Norouzi, Mohammad and Hinton, Geoffrey},
  journal={arXiv preprint arXiv:2002.05709},
  year={2020}
}

@InProceedings{alp,
author="Cicimen, Alp G.
and Tregidgo, Henry F. J.
and Figini, Matteo
and Messaritaki, Eirini
and B. McNabb, Carolyn
and Palombo, Marco
and Evans, C. John
and Cercignani, Mara
and Jones, Derek K.
and C. Alexander, Daniel",
title="Image Quality Transfer of Diffusion MRI Guided By High-Resolution Structural MRI",
booktitle="Computational Diffusion MRI",
year="2025",
}

@article{flowchef,
        title={Steering Rectified Flow Models in the Vector Field for Controlled Image Generation},
        author={Patel, Maitreya and Wen, Song and Metaxas, Dimitris N. and Yang, Yezhou},
        journal={arXiv preprint arXiv:2412.00100},
        year={2024}
      }

@inproceedings{dpps,
  title={Diffusion Posterior Proximal Sampling for Image Restoration},
  author={Wu, Hongjie and He, Linchao and Zhang, Mingqin and Chen, Dongdong and Luo, Kunming and Luo, Mengting and Zhou, Ji-Zhe and Chen, Hu and Lv, Jiancheng},
  booktitle={Proceedings of the 32nd ACM International Conference on Multimedia},
  year={2024}
}

@article{resnet,
	author = {Kaiming He and Xiangyu Zhang and Shaoqing Ren and Jian Sun},
	title = {Deep Residual Learning for Image Recognition},
	journal = {arXiv preprint arXiv:1512.03385},
	year = {2015}
}

@misc{flowdps,
      title={FlowDPS: Flow-Driven Posterior Sampling for Inverse Problems}, 
      author={Jeongsol Kim and Bryan Sangwoo Kim and Jong Chul Ye},
      year={2025},
      eprint={2503.08136},
      archivePrefix={arXiv},
}

@article{rectifiedflow,
  title={Flow straight and fast: Learning to generate and transfer data with rectified flow},
  author={Liu, Xingchao and Gong, Chengyue and Liu, Qiang},
  journal={arXiv preprint arXiv:2209.03003},
  year={2022}
}

@InProceedings{swinunet,
author = {Hu Cao and Yueyue Wang and Joy Chen and Dongsheng Jiang and Xiaopeng Zhang and Qi Tian and Manning Wang},
title = {Swin-Unet: Unet-like Pure Transformer for Medical Image Segmentation},
booktitle = {Proceedings of the European Conference on Computer Vision Workshops(ECCVW)},
year = {2022}
}
%




\end{document}